\documentclass{article}
\usepackage{main,times}

\usepackage{amsmath}

\usepackage[T1]{fontenc}
\usepackage{amssymb}
\usepackage{graphicx}
\usepackage{booktabs}
\usepackage{multirow}
\usepackage{xcolor}
\usepackage{microtype}
\usepackage[hidelinks]{hyperref}
\usepackage{url}
\usepackage{array}
\usepackage{wrapfig}
\usepackage[capitalise,nameinlink]{cleveref}

\Crefformat{section}{\S#2#1#3}
\Crefformat{subsection}{\S#2#1#3}
\Crefrangeformat{section}{\S#3#1#4 to~\S#5#2#6}
\Crefrangeformat{subsection}{\S#3#1#4 to~\S#5#2#6}
\Crefmultiformat{section}{\S#2#1#3}{ and~\S#2#1#3}{, \S#2#1#3}{ and~\S#2#1#3}
\Crefmultiformat{subsection}{\S#2#1#3}{ and~\S#2#1#3}{, \S#2#1#3}{ and~\S#2#1#3}
\Crefformat{appendix}{Appx.~#2#1#3}
\Crefmultiformat{appendix}{Appx.~#2#1#3}{ and~#2#1#3}{, #2#1#3}{ and~#2#1#3}
\Crefformat{figure}{#2Fig.~#1#3}
\Crefmultiformat{figure}{Figs.~#2#1#3}{ and~#2#1#3}{, #2#1#3}{ and~#2#1#3}
\Crefformat{table}{#2Tab.~#1#3}
\Crefmultiformat{table}{Tabs.~#2#1#3}{ and~#2#1#3}{, #2#1#3}{ and~#2#1#3}

\definecolor{gainc}{HTML}{1B7F5A}
\definecolor{lossc}{HTML}{B3402F}
\definecolor{srcswap}{HTML}{A8781F}
\definecolor{srccross}{HTML}{C2622A}
\definecolor{srcoff}{HTML}{A03123}
\definecolor{refline}{HTML}{6E6A63}
\definecolor{bcliberopro10swap}{HTML}{2F6FB5}
\definecolor{bcrobocerebrapublic60}{HTML}{7A4FA3}
\definecolor{bcvlabenchxvlatracks14}{HTML}{C2622A}
\definecolor{bcsimplerwidowxvm}{HTML}{17868A}
\definecolor{bcsimplergooglevm}{HTML}{9A7B1F}
\definecolor{bcrobocasa365target}{HTML}{2E7D4F}
\newcommand{\method}{\textsc{RoboArchitect}}
\newcommand{\para}[2][]{\smallskip\noindent
  \ifx\relax#1\relax\else\phantomsection\label{#1}\fi
  \textbf{#2}\hspace{0.4em}}
\newcommand{\panelcap}[1]{\parbox{\textwidth}{\raggedright\scriptsize #1}}

\newcommand{\bench}[2]{#2}
\newcommand{\lib}{\bench{bcliberopro10swap}{LIBERO-Pro-10}}
\newcommand{\cere}{\bench{bcrobocerebrapublic60}{RoboCerebra}}
\newcommand{\vlab}{\bench{bcvlabenchxvlatracks14}{VLABench}}
\newcommand{\widow}{\bench{bcsimplerwidowxvm}{SimplerEnv WidowX}}
\newcommand{\goog}{\bench{bcsimplergooglevm}{SimplerEnv Google}}
\newcommand{\casa}{\bench{bcrobocasa365target}{RoboCasa-365}}

\title{\method: Automating the Design of\\Robotic Agents with Persistent Memory}

\author{Shengguang Wu\hspace{0.6em} Suning Huang\hspace{0.6em} Cherie Ho\hspace{0.6em} Ria Doshi\hspace{0.6em} Hao Zhu\hspace{0.6em} Serena Yeung-Levy \\
\normalfont Stanford University \\[4pt]
\normalfont \url{https://roboarchitect-ai.github.io/}}

\iclrfinalcopy
\begin{document}

\maketitle
\lhead{Preprint}
\suppressfloats[t]

\begin{abstract}
Recent work shows that a frozen robot policy performs far better inside an agent harness, where control code decides when to query the policy, consult vision tools, or command the arm directly. 
Yet today these harnesses are built by hand and redesigned for each new robot or benchmark.
We show that harness design can be automated, offering a strong alternative to finetuning on a novel setup.
In \method{}, a coding model observes a robot's rollouts, diagnoses failures, and iteratively revises the harness around the frozen policy.
The harness evolves on a set of episodes for its optimization and is evaluated on held-out scenes and tasks.
Without updating any model weights, \method{} raises average success from 9.7\% to 43.3\% on unseen scenes (4.5$\times$) and from 12.7\% to 40.2\% on unseen tasks (3.2$\times$), across multiple benchmarks and policies.
The gains are largest where the benchmark is far from the policy's training data: wrapping a general-purpose policy, the evolved harness yields a specialist that often outperforms even the finetuned version.
Intuitive control strategies are automatically designed, such as lifting a jammed arm clear or reaching in horizontally when an obstacle blocks the way from above.
The harness also evolves its own persistent memory, carrying learned experience, such as effective grasp depths, into later decisions.
These results suggest that substantial headroom in robotic manipulation lies in the agent harness, and its design can be effectively automated and optimized without weight updates.
\end{abstract}

\section{Introduction} \label{sec:intro}

Language models have become far more useful inside an agent harness: surrounding code that mediates when and how to call the model, what context to supply, and how to handle its output.
Robot policies are undergoing the same shift. 
This shift is particularly important for robotics, where adapting a policy to new tasks or hardware often requires collecting new demonstrations and finetuning model weights~\citep{kim2024openvla,zang2025rlinf,park2026retrieve}.
We posit that \textbf{harness optimization offers an effective alternative to finetuning, turning a general policy into a specialist} for a specific robot setup (\cref{sec:headroom}).

Recent agentic robot systems like VoLo~\citep{chen2026volo} and Harness VLA~\citep{zhang2026harness} show strong gains from wrapping frozen policies in hand-written harnesses~\citep{xu2026don,zhao2026skills,gao2026agm,hu2026matters}.
Today, these harnesses are still manually designed, often requiring substantial engineering effort for each task and embodiment.
Building such a harness follows a familiar trial-and-error loop: run the agent on the robot, observe where it fails, diagnose the cause, modify the control code, and test again.
This work is slow to perform by hand, and long recordings of robot rollouts are difficult for a human engineer to inspect exhaustively.
We ask \textbf{whether this complete engineering loop, from diagnosing failure traces to refining the harness design to fix them, can be automated}.

In this work, \method{} uses a coding model to automatically design and optimize the harness ({\cref{sec:method}}).
Given access to a frozen policy, auxiliary vision and language tools, and step-by-step rollout recordings, the coding model diagnoses failure modes and proposes an updated harness to address them.
Like a finetuned policy, a harness optimized on specific rollouts could overfit to them.
We therefore split each benchmark into ``training'' (for harness optimization) and held-out episodes, evaluating on \emph{unseen scenes}, where tasks are shared but scene arrangements differ, and \emph{unseen tasks}, where held-out episodes contain entirely different tasks that require shared manipulation skills (\cref{para:protocols}).

\begin{wrapfigure}{R}{0.63\textwidth}
 \centering
 \includegraphics[width=0.63\textwidth]{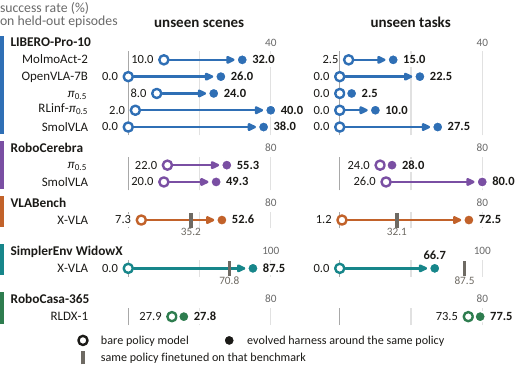}
 \caption{\textbf{Wrapping a frozen policy in an evolved harness raises its held-out success rate from 9.7\% to 43.3\% on unseen scenes and from 12.7\% to 40.2\% on unseen tasks, often past the same policy finetuned on the benchmark} (\cref{para:specialist}). All scores are evaluated on the held-out episodes never seen during the harness's optimization. \emph{Unseen scenes:} held-out episodes evaluate the same tasks under varied scene arrangements. \emph{Unseen tasks:} held-out episodes evaluate distinct tasks to test cross-task skill transfer (\cref{para:protocols}).}
 \label{fig:teaser}
\end{wrapfigure}
Across diverse benchmarks and policies, our auto-optimized harnesses raise the average success rate on unseen scenes from 9.7\% to 43.3\% (4.5$\times$), and on unseen tasks from 12.7\% to 40.2\% (3.2$\times$) (\cref{fig:teaser}).
The strongest gains occur when policies are deployed in out-of-domain settings, where \method{} can effectively substitute for new training data: wrapping a general policy yields a specialist that often outperforms the finetuned version of the same model.
For example, in settings where the vision-language-action (VLA) policy contributes no useful signal, the harness learns to stop querying it and instead drives the arm from its own perception, effectively becoming a code-as-policy agent.
While code-as-policy~\citep{liang2023code,lu2026aspire,elmaaroufi2026rho} and learned VLA policies~\citep{kim2024openvla,intelligence2025pi_} have traditionally been treated as distinct paradigms, \method{} unifies them empirically (\cref{para:balance}).
Unlike weight updates, each harness revision is also fully interpretable: it targets specific failure modes with intuitive control logic (\cref{fig:drivers1,fig:drivers2}), making subsequent revisions easy to inspect and build upon, and such iterative refinement proves to be key to generalization (\cref{para:iterative}).

We also equip the harness with \textbf{persistent memory}, allowing experience to accumulate across episodes. No schema or memory rules are specified; \method{} is entirely free to decide what to store, in what form, and how to retrieve it.
As a result, diverse forms of memories emerge from the harness optimization to record key pragmatic clues, such as effective grasp depths that inform later control decisions.

As \method{} proves highly effective in solving tasks on one robot embodiment at a time, we further use its harness evolution to build \textbf{a unified, generalizable agent that operates across different robot platforms} (\cref{para:unified}).
In this single harness optimized on a training pool drawn from diverse benchmarks, shared control strategies and memory structures emerge naturally across embodiments.

Our findings suggest that substantial performance headroom in robotic manipulation lies in the agent harness, as an alternative to finetuning. 
\method{} automates its design and optimization, providing a practical way to adapt robot policies to new environments without updating weights.

\section{Related Work} \label{sec:related}

\para{Vision-language-action policies.} 
VLA policies are pretrained vision-language models finetuned for robot control, mapping visual observations and language instructions directly to robot actions~\citep{brohan2023rt,kim2024openvla,black2024pi_0,intelligence2025pi_,shukor2025smolvla,zheng2026x,fang2026molmoact2,kim2026rldx}. 
In practice, their performance on new, out-of-distribution tasks can depend heavily on finetuning~\citep{zang2025rlinf,park2026retrieve}, which requires collecting demonstrations for every new task or hardware configuration.
Such finetuned manipulation skills often fail to generalize beyond initial training distributions~\citep{zhou2025libero}.
Our held-out protocols test exactly this transfer (\cref{para:protocols}), on which the evolved harnesses maintain substantially better performance (\cref{sec:headroom}).
A complementary line of work keeps the policy frozen and adds inference-time mechanisms: retrieving demonstrations or successful action segments to condition the policy~\citep{park2026retrieve,zhao2026retrieve}, or selecting among the policy's candidate actions with a learned value function~\citep{xie2026look,zhang2026frozen}.
In contrast, \method{} \textbf{evolves complete agent interfaces from scratch that combine various mechanisms}, including persistent memory, programmatic control, and policy orchestration, all custom-fit to each setup (\cref{sec:changes,sec:memory-results}).

\para{Robotic agents.} 
Code-as-policy methods~\citep{liang2023code,singh2022progprompt,huang2023voxposer} bypass trained policies, and instead use executable programs as the primary robot control, composing perception, reasoning, and motion primitives.
Recent coding agents go further, using iterative code generation to solve manipulation tasks~\citep{tsui2026free,kumar2026act,xie2026revisiting}. 
However, CaP-X~\citep{fu2026cap} shows that their success depends heavily on the hand-written primitives they are given.
Other recent work retains trained VLAs but places them inside agent-level orchestration. 
VoLo~\citep{chen2026volo} treats the policy as an interruptible tool, guiding it alongside vision models and programmatic motions. 
Harness VLA~\citep{zhang2026harness} assigns high-level planning to an LLM agent and restricts the frozen policy to contact-level primitives. 
Several systems further decompose tasks into subproblems with memory or verification mechanisms~\citep{xu2026don,gao2026agm,zhao2026skills}.
\citet{hu2026matters} find that components in such hierarchical VLA architectures (planner, controllers, and switching logic) must be co-designed to maximize performance.
These works establish that the interface around a VLA matters, but their agent architectures were specified manually in advance.
\method{} instead automates harness design and optimization, including the VLA/code division of labor (\cref{para:balance}) and persistent memory (\cref{sec:memory-results}).

\para{Automated code optimization.}
Coding models are increasingly used in auto-research routines that evaluate agent outputs, analyze execution traces, and iteratively refine agent implementations~\citep{hu2025automated,zhang2026darwin,lee2026meta,zhang2026self,wei2026evo,zhou2026hierarchical,cai2026moss}.
Within robotics, most LLM-driven optimization loops generate artifacts that train policy weights: reward functions (Eureka~\citep{ma2024eureka}), training configurations revised from execution logs (ENPIRE~\citep{xiao2026enpire}), or VLM-decomposed exploration data distilled into policies via finetuning (EXIMO~\citep{sukhija2026eximo}).
These systems \textbf{automate finetuning}, and are thus complementary to \method{}, which \textbf{automates the agent built around a frozen policy}.
Another line of work accumulates reusable code skills: repairing and archiving control functions~\citep{lu2026aspire}, searching over repositories of control code~\citep{elmaaroufi2026rho}, assembling computation graphs from skill libraries~\citep{chen2026gap}, collecting skills during self-directed play~\citep{zhang2026playful}, or distilling language corrections into a skill library for in-context learning~\citep{chen2026few}.
Crucially, these skill systems optimize programs that \textbf{entirely replace the robot policy}.
In contrast, \method{} develops the harness \emph{around} a frozen policy, allowing the code to \textbf{autonomously divide control between the policy and its own direct commands} (\cref{para:balance}).
\citet{claudeplaysrobotics2026} similarly find that frontier coding models are more capable when given higher-level access to pretrained robot policies than when controlling low-level motion directly, an arrangement \method{} often evolves into on its own.

\begin{figure}[t]
 \centering
 \includegraphics[width=\textwidth]{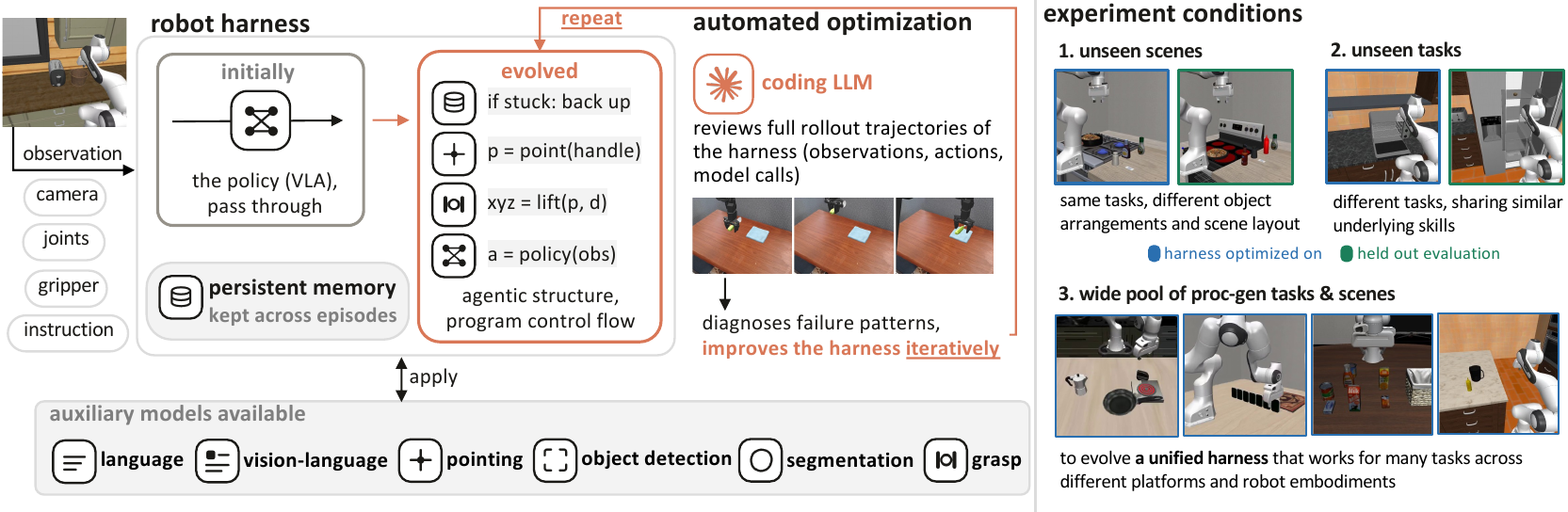}
\caption{\textbf{\method{} turns a frozen policy into a performant robotic agent without updating its weights: a coding model iteratively refines the harness from execution traces.} Starting from a bare policy (VLA), \method{} evolves a robotic agent equipped with auxiliary tools and persistent memory. The right side illustrates experiment conditions (\cref{para:protocols}) that test how well the evolved harness generalizes.}
 \label{fig:method}
\end{figure}

\section{\method{}} \label{sec:method}

\method{} optimizes the harness around a frozen policy, as illustrated in \cref{fig:method}.
It starts from a harness that passes the policy's actions through unchanged and repeats an empirical optimization loop: the harness runs on the training episodes, a coding model observes the recorded rollouts, diagnoses the failures and refines the harness.
These ``training'' episodes serve only to optimize the harness: \textbf{no weights are updated}, and the final harness is evaluated on separate held-out episodes (\cref{para:protocols}).
Over iterations, the harness grows into a highly performant robotic agent that, as needed, follows the policy, consults auxiliary tools, or commands the arm directly, and it can carry experience in a persistent memory.
Below, we describe its key components.

\para{Harness interface.}
A robotic agent harness is a program that sits between environment observations and low-level robot execution commands.
At each control step, the harness receives the current environment observation, including camera images, depth maps, gripper pose, and gripper status, along with the natural language task instruction.
It outputs low-level action vectors executed by the robot controller.
Initially, the harness is a simple pass-through of the bare VLA policy: it forwards observations to the frozen policy and returns the policy's predicted actions directly.
The \emph{Bare Policy} columns of \cref{tab:main} are thus the standard evaluation of the policy alone.

\para{Auxiliary tools.}
Beyond the base policy, the harness can query auxiliary pretrained tools (\cref{para:config}), including language models, vision-language models, object detectors, segmentation models, pointing models, and grasp pose generators.
The harness also has access to geometric utilities that project pixel-depth pairs into 3D robot base coordinates, and generate relative trajectory primitives.
Together, these tools give the harness alternatives to relying on the frozen policy alone.

\para[para:optimizationloop]{Optimization loop.}
The optimization follows the same trial-and-error loop as a human engineer (\cref{sec:intro}).
Every evaluation logs complete trajectory traces, including per-step camera frames, the robot's readings, and all model and tool outputs.
A coding model reviews these execution traces from the current harness, diagnoses failure modes, and proposes a candidate harness revision, where it may follow the policy, invoke any of the tools, or command the arm programmatically.
The automated optimization loop thus settles the balance between policy and programmatic control (\cref{para:balance}).
The candidate is then re-evaluated on the same training set (\cref{para:protocols}) and accepted only if its performance surpasses that of the current harness, following the empirical validation in prior work on self-improving agents~\citep{zhang2026darwin,zhang2026self}.
We thus ensure a monotonic optimization trajectory on the training set, capped at a maximum number of iterations (\cref{para:config}). We compare iterative refinement with a single-attempt automated agent design in \cref{para:iterative}.

\para{Persistent memory.}
We also provide the harness with a persistent memory mechanism that carries information across episodes and tasks.
Unlike fixed hand-crafted memory schemes in recent robotic agents~\citep{zhang2026harness,xu2026don,zhao2026skills,gao2026agm}, our persistent memory structure is \textbf{autonomously evolved based on accumulated rollout evidence}, and fitted to each setup.
As discussed in \cref{sec:memory-results}, \method{} discovers diverse forms of persistent memory.

\section{Experiments} \label{sec:experiments}

\subsection{Setup} \label{sec:setup}

\para{Benchmarks and policies.}
We evaluate on \lib{}, which is LIBERO-10 (\emph{i.e.,} LIBERO-Long)~\citep{liu2023libero} with object locations swapped as in LIBERO-Pro~\citep{zhou2025libero}, \cere{}~\citep{han2025robocerebra}, \vlab{}~\citep{zhang2025vlabench}, \casa{}~\citep{nasiriany2024robocasa,nasiriany2026robocasa365} and \widow{}~\citep{li2024evaluating}. The unified harness (\cref{para:protocols}) is also tested on \goog{}~\citep{li2024evaluating}. \cref{para:appendix-bench} describes each benchmark.
The VLA policies are MolmoAct-2~\citep{fang2026molmoact2}, OpenVLA-7B~\citep{kim2024openvla}, $\pi_{0.5}$~\citep{intelligence2025pi_}, RLinf-$\pi_{0.5}$~\citep{chen2025pi_,zang2025rlinf}, SmolVLA~\citep{shukor2025smolvla}, X-VLA~\citep{zheng2026x}, and RLDX-1~\citep{kim2026rldx}.

\para[para:distance]{Training-source distance.}
We characterize each setup by the distance between the policy's training data and the evaluation benchmark (\cref{tab:main}): (1)~in-distribution, where the policy was tuned on this exact benchmark; (2)~\textcolor{srcswap}{swapped}, with object locations permuted as in LIBERO-Pro; (3)~\textcolor{srccross}{cross-benchmark}, a different benchmark on the same simulator and robot; and (4)~\textcolor{srcoff}{out-of-domain}, with only general robot pretraining and benchmark unseen.
For comparison with finetuning, we also evaluate public checkpoints of the same policy finetuned on the target benchmark on the same held-out test set (rows marked $\hookrightarrow$ in \cref{tab:main}).

\para[para:protocols]{Held-out evaluation protocols.}
Every harness is evaluated on held-out episodes never seen during the harness's optimization (\emph{i.e.,} ``training''), under the following protocols: 

(1) \textbf{\emph{Unseen scenes}}.
Training and held-out splits share identical tasks but differ in initial scene arrangements and layouts (\cref{fig:splits}, left). The evolved harness must generalize beyond specific object positions or spatial configurations.

(2) \textbf{\emph{Unseen tasks}}.
Training and held-out splits contain different tasks that share underlying manipulation skills (\cref{fig:splits}, right). The evolved harness must generalize at the skill level rather than memorizing task-specific routines.

(3) \textbf{\emph{Unified harness across benchmarks}}.
Under the protocols above, individual agents are evolved separately for each setup, allowing them to specialize to the challenges of specific embodiments.
As an extension, we also evolve a \textbf{single unified harness that works across multiple benchmarks simultaneously}, spanning different robot embodiments, simulators and action layouts, with the goal of discovering shared cross-embodiment agent mechanisms and persistent memory structures.
We take the benchmarks LIBERO-Pro, \vlab{}, \widow{} and \goog{} as test bed (Franka Panda, WidowX and the Google mobile manipulator).
We use the following policy configurations: (1)~the \emph{general} X-VLA checkpoint, never tuned on any of these benchmarks and thus operating entirely out of distribution; (2)~\emph{X-VLA specialists}, one checkpoint finetuned per benchmark.
To evolve the unified harness, we build a training pool of \textbf{procedurally generated episodes with new scene arrangements, objects, instructions or tasks} outside standard benchmark evaluations (\cref{fig:unified-pool}, top), and evaluate on the benchmarks' own standard episodes (\cref{fig:unified-pool}, bottom).

\para{Evaluation metrics.}
We report standard benchmark metrics. On \vlab{}, we use the official progress score, which assigns partial credit for grasping the right object and for each goal condition met; on all other benchmarks, we report success rate averaged over tasks. For simplicity, we call all these scores success rates (\%) in this paper, although the \vlab{} numbers are progress scores. 
All the scores we compare, including those of the finetuned references, use the same step limits. On \widow{} and \goog{}, episodes always run to the limit and are scored on their final state.

\para[para:config]{Harness evolution configurations.}
We run at most 5 optimization iterations, using \texttt{claude-opus-5} as the coding model, invoked via the Claude Code client at maximum effort.
The \textbf{auxiliary tools} are Qwen3-30B-A3B-Instruct~\citep{yang2025qwen3} as the language model, Molmo2-8B~\citep{clark2026molmo2} or Qwen3-VL-8B-Instruct~\citep{bai2025qwen3} as the vision-language model, Grounding-DINO~\citep{liu2024grounding} for object detection, SAM~3~\citep{carion2026sam} for segmentation, Molmo2-8B for pointing, and GraspGen~\citep{murali2025graspgen} for grasp pose generation.
Further implementation details are provided in \cref{sec:appendix-impl}.

\begin{table}[t]
 \centering
\caption{\textbf{\method{}'s iteratively evolved harness raises mean held-out success by +345\% on unseen scenes and +216\% on unseen tasks.} \emph{Bare Policy}: the VLA policy alone. \emph{Harness (ours)}: the same frozen policy inside the harness \method{} designed, after a single design attempt (\emph{One-Shot}, iteration 1) or after iterative refinement over up to 5 iterations (\emph{Iterative}). $\Delta$: gain of the iterative harness over the bare policy. Rows marked $\hookrightarrow$ show the finetuned version of the same policy evaluated on the same held-out set. Training-source distance between the policy and the evaluation benchmark is color-coded: black = in-distribution; \textcolor{srcswap}{gold} = spatially swapped; \textcolor{srccross}{orange} = cross-benchmark; \textcolor{srcoff}{dark red} = out-of-domain. All scores are success rates in \%.}
 \label{tab:main}
 \scriptsize
 \setlength{\tabcolsep}{2.5pt}
 \begin{tabular}{@{}lllrr r@{\hspace{\dimexpr2\tabcolsep+3pt}}r@{\hspace{\dimexpr2\tabcolsep+7pt}}rr r@{\hspace{\dimexpr2\tabcolsep+3pt}}r@{}}
 \toprule
& & & \multicolumn{4}{c}{Unseen Scenes (\%)} & \multicolumn{4}{c}{Unseen Tasks (\%)} \\
 \cmidrule(lr){4-7}\cmidrule(lr){8-11}
& & & \multicolumn{1}{c}{Bare} & \multicolumn{2}{c}{Harness (ours)} & \multicolumn{1}{c}{\multirow{2}{*}{$\Delta$}} & \multicolumn{1}{c}{Bare} & \multicolumn{2}{c}{Harness (ours)} & \multicolumn{1}{c}{\multirow{2}{*}{$\Delta$}} \\
 \cmidrule(lr){5-6}\cmidrule(lr){9-10}
Benchmark & Policy & Trained On & \multicolumn{1}{c}{Policy} & \multicolumn{1}{c}{One-Shot} & \multicolumn{1}{c}{Iterative} & & \multicolumn{1}{c}{Policy} & \multicolumn{1}{c}{One-Shot} & \multicolumn{1}{c}{Iterative} & \\
 \midrule
\textcolor{bcliberopro10swap}{\textbf{LIBERO-Pro-10}} & MolmoAct-2 & {\scriptsize \textcolor{srcswap}{LIBERO}} & 10.0 & 12.0 & \textbf{32.0} & \textcolor{gainc}{+22.0} & 2.5 & 2.5 & \textbf{15.0} & \textcolor{gainc}{+12.5} \\
 & OpenVLA-7B & {\scriptsize \textcolor{srcswap}{LIBERO-10}} & 0.0 & 2.0 & \textbf{26.0} & \textcolor{gainc}{+26.0} & 0.0 & 0.0 & \textbf{22.5} & \textcolor{gainc}{+22.5} \\
 & $\pi_{0.5}$ & {\scriptsize \textcolor{srcswap}{LIBERO}} & 8.0 & 4.0 & \textbf{24.0} & \textcolor{gainc}{+16.0} & 0.0 & 0.0 & \textbf{2.5} & \textcolor{gainc}{+2.5} \\
 & RLinf-$\pi_{0.5}$ & {\scriptsize \textcolor{srcswap}{LIBERO-130}} & 2.0 & 2.0 & \textbf{40.0} & \textcolor{gainc}{+38.0} & 0.0 & 0.0 & \textbf{10.0} & \textcolor{gainc}{+10.0} \\
 & SmolVLA & {\scriptsize \textcolor{srccross}{RoboCerebra}} & 0.0 & 0.0 & \textbf{38.0} & \textcolor{gainc}{+38.0} & 0.0 & 10.0 & \textbf{27.5} & \textcolor{gainc}{+27.5} \\
\textcolor{bcrobocerebrapublic60}{\textbf{RoboCerebra}} & $\pi_{0.5}$ & {\scriptsize \textcolor{srccross}{LIBERO}} & 22.0 & 20.0 & \textbf{55.3} & \textcolor{gainc}{+33.3} & 24.0 & 12.0 & \textbf{28.0} & \textcolor{gainc}{+4.0} \\
 & SmolVLA & {\scriptsize RoboCerebra} & 20.0 & 34.3 & \textbf{49.3} & \textcolor{gainc}{+29.3} & 26.0 & 16.0 & \textbf{80.0} & \textcolor{gainc}{+54.0} \\
\textcolor{bcvlabenchxvlatracks14}{\textbf{VLABench}} & X-VLA & {\scriptsize \textcolor{srcoff}{general robot data}} & 7.3 & 32.8 & \textbf{52.6} & \textcolor{gainc}{+45.2} & 1.2 & -- & \textbf{72.5} & \textcolor{gainc}{+71.3} \\
\multicolumn{3}{@{}l}{\hspace{1em}\textcolor{refline}{$\hookrightarrow$ \emph{X-VLA finetuned on VLABench}}} & \textcolor{refline}{35.2} & \textcolor{refline}{--} & \textcolor{refline}{--} & \textcolor{refline}{--} & \textcolor{refline}{32.1} & \textcolor{refline}{--} & \textcolor{refline}{--} & \textcolor{refline}{--} \\
\textcolor{bcsimplerwidowxvm}{\textbf{SimplerEnv WidowX}} & X-VLA & {\scriptsize \textcolor{srcoff}{general robot data}} & 0.0 & 77.1 & \textbf{87.5} & \textcolor{gainc}{+87.5} & 0.0 & -- & \textbf{66.7} & \textcolor{gainc}{+66.7} \\
\multicolumn{3}{@{}l}{\hspace{1em}\textcolor{refline}{$\hookrightarrow$ \emph{X-VLA finetuned on WidowX}}} & \textcolor{refline}{70.8} & \textcolor{refline}{--} & \textcolor{refline}{--} & \textcolor{refline}{--} & \textcolor{refline}{87.5} & \textcolor{refline}{--} & \textcolor{refline}{--} & \textcolor{refline}{--} \\
\textcolor{bcrobocasa365target}{\textbf{RoboCasa-365}} & RLDX-1 & {\scriptsize RoboCasa-365} & 27.9 & 29.3 & \textbf{27.8} & \textcolor{lossc}{-0.2} & 73.5 & 76.5 & \textbf{77.5} & \textcolor{gainc}{+4.0} \\
\midrule
\multicolumn{3}{@{}l}{\textit{Mean (all setups)}} & 9.7 & 21.4 & \textbf{43.3} & \textcolor{gainc}{+345\%} & 12.7 & -- & \textbf{40.2} & \textcolor{gainc}{+216\%} \\
\multicolumn{3}{@{}l}{\textit{Mean (cross-benchmark and out-of-domain)}} & 7.3 & 32.5 & \textbf{58.4} & \textcolor{gainc}{+696\%} & 6.3 & -- & \textbf{48.7} & \textcolor{gainc}{+671\%} \\
\bottomrule
 \end{tabular}
\end{table}

\begin{table}[t]
 \centering
\caption{\textbf{A unified evolved harness serves different benchmarks: around the general X-VLA it substantially raises held-out success, and around the finetuned X-VLA specialists it preserves their competence.} Columns and labels read as in \cref{tab:main}.}
 \label{tab:unified}
 \scriptsize
 \setlength{\tabcolsep}{2.5pt}
 \begin{tabular}{@{}lllrr r@{\hspace{\dimexpr2\tabcolsep+3pt}}r@{\hspace{\dimexpr2\tabcolsep+7pt}}rr r@{\hspace{\dimexpr2\tabcolsep+3pt}}r@{}}
 \toprule
& & & \multicolumn{4}{c}{\shortstack{Training Pool (\%)\\{\scriptsize procedurally generated}}} & \multicolumn{4}{c}{\shortstack{Held-Out (\%)\\{\scriptsize standard benchmark episodes}}} \\
 \cmidrule(lr){4-7}\cmidrule(lr){8-11}
& & & \multicolumn{1}{c}{Bare} & \multicolumn{2}{c}{Harness (ours)} & \multicolumn{1}{c}{\multirow{2}{*}{$\Delta$}} & \multicolumn{1}{c}{Bare} & \multicolumn{2}{c}{Harness (ours)} & \multicolumn{1}{c}{\multirow{2}{*}{$\Delta$}} \\
 \cmidrule(lr){5-6}\cmidrule(lr){9-10}
Policy & Trained On & Benchmark & \multicolumn{1}{c}{Policy} & \multicolumn{1}{c}{One-Shot} & \multicolumn{1}{c}{Iterative} & & \multicolumn{1}{c}{Policy} & \multicolumn{1}{c}{One-Shot} & \multicolumn{1}{c}{Iterative} & \\
 \midrule
\multirow{5}{*}{\shortstack[l]{General\\X-VLA}} & \multirow{5}{*}{{\scriptsize \textcolor{srcoff}{\shortstack[l]{general\\robot data}}}} & \textcolor{bcliberopro10swap}{\textbf{LIBERO-Pro}} & 0.0 & 3.5 & \textbf{29.4} & \textcolor{gainc}{+29.4} & 0.1 & 3.5 & \textbf{24.1} & \textcolor{gainc}{+24.0} \\
 &  & \textcolor{bcvlabenchxvlatracks14}{\textbf{VLABench}} & 7.1 & 14.8 & \textbf{17.5} & \textcolor{gainc}{+10.4} & 7.5 & 17.3 & \textbf{21.6} & \textcolor{gainc}{+14.2} \\
 &  & \textcolor{bcsimplerwidowxvm}{\textbf{SimplerEnv WidowX}} & 0.0 & 65.8 & \textbf{68.3} & \textcolor{gainc}{+68.3} & 0.0 & 45.8 & \textbf{24.0} & \textcolor{gainc}{+24.0} \\
 &  & \textcolor{bcsimplergooglevm}{\textbf{SimplerEnv Google}} & 0.0 & 22.0 & \textbf{51.2} & \textcolor{gainc}{+51.2} & 0.0 & 20.0 & \textbf{45.6} & \textcolor{gainc}{+45.6} \\
 \cmidrule(lr){3-11}
 &  & \textit{Mean over all benchmarks} & 1.8 & 26.5 & \textbf{41.6} & \textcolor{gainc}{+2228\%} & 1.9 & 21.6 & \textbf{28.8} & \textcolor{gainc}{+1417\%} \\
\midrule
\multirow{5}{*}{\shortstack[l]{Specialist\\X-VLA}} & \multirow{5}{*}{{\scriptsize \shortstack[l]{finetuned on\\each benchmark}}} & \textcolor{bcliberopro10swap}{\textbf{LIBERO-Pro}} & 82.9 & 83.2 & \textbf{82.8} & \textcolor{lossc}{-0.1} & 8.4 & 8.1 & \textbf{8.0} & \textcolor{lossc}{-0.4} \\
 &  & \textcolor{bcvlabenchxvlatracks14}{\textbf{VLABench}} & 13.4 & 14.9 & \textbf{23.6} & \textcolor{gainc}{+10.2} & 30.4 & 33.5 & \textbf{35.1} & \textcolor{gainc}{+4.7} \\
 &  & \textcolor{bcsimplerwidowxvm}{\textbf{SimplerEnv WidowX}} & 80.0 & 80.0 & \textbf{80.8} & \textcolor{gainc}{+0.8} & 72.9 & 72.9 & \textbf{74.0} & \textcolor{gainc}{+1.0} \\
 &  & \textcolor{bcsimplergooglevm}{\textbf{SimplerEnv Google}} & 55.9 & 55.9 & \textbf{55.4} & \textcolor{lossc}{-0.4} & 71.9 & 71.7 & \textbf{70.9} & \textcolor{lossc}{-1.0} \\
 \cmidrule(lr){3-11}
 &  & \textit{Mean over all benchmarks} & 58.0 & 58.5 & \textbf{60.7} & \textcolor{gainc}{+5\%} & 45.9 & 46.5 & \textbf{47.0} & \textcolor{gainc}{+2\%} \\
\bottomrule
 \end{tabular}
\end{table}

\begin{figure}[t]
 \centering
 \includegraphics[width=\textwidth]{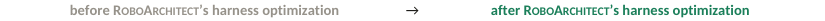}\\[2pt]
 \includegraphics[width=\textwidth]{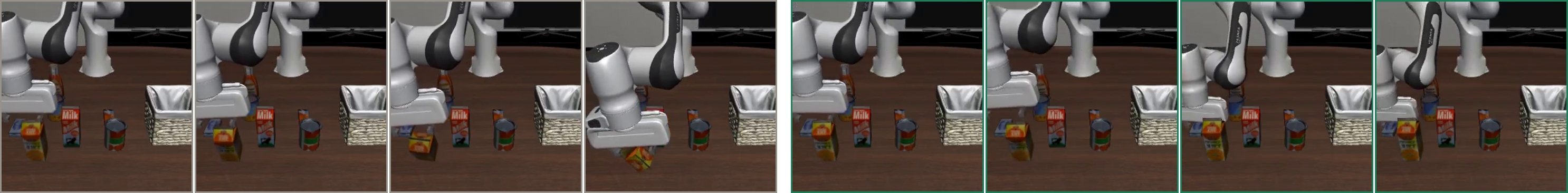}\\[1.5pt]
\panelcap{\textbf{(a)}~MolmoAct-2 on \lib{}, \emph{``put both the cream cheese box and the butter in the basket''}: the VLA policy keeps driving into the juice carton until it tips over; the evolved harness detects the jam and lifts clear.}\\[5pt]
 \includegraphics[width=\textwidth]{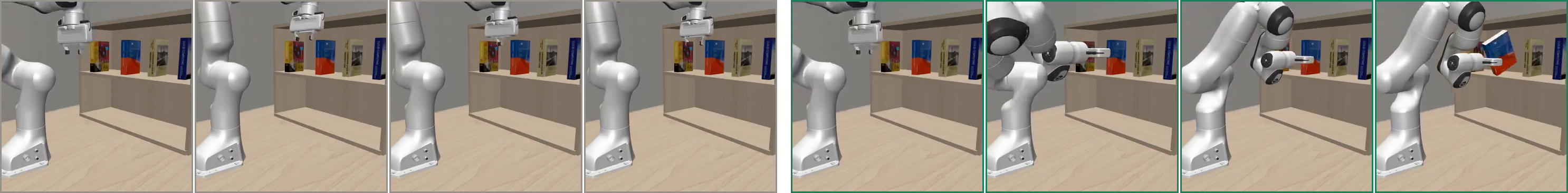}\\[1.5pt]
\panelcap{\textbf{(b)}~X-VLA on \vlab{}, \emph{``Please take the book contract\_law\_in\_japan''}: the earlier harness defaults to top-down motion and collides repeatedly with the shelf board from above; the evolved harness turns the wrist and reaches in from the side.}\\[5pt]
 \includegraphics[width=\textwidth]{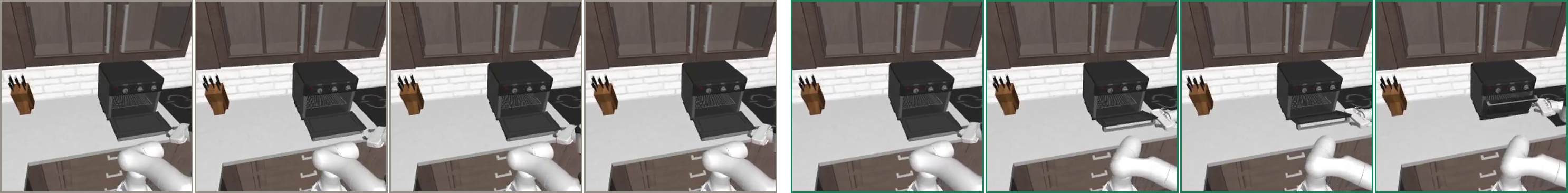}\\[1.5pt]
\panelcap{\textbf{(c)}~RLDX-1 on \casa{}, \emph{``Close the toaster oven door''}: the VLA policy leans on the toaster-oven door too gently and it stays open; the evolved harness rescales the action magnitude and latches the door shut.}\\[5pt]
 \includegraphics[width=\textwidth]{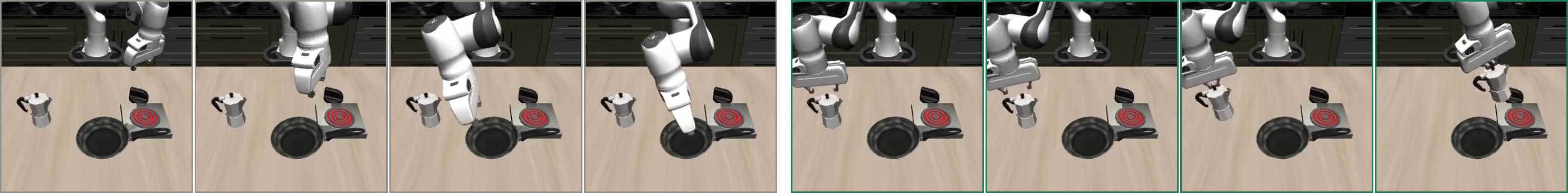}\\[1.5pt]
\panelcap{\textbf{(d)}~RLinf-$\pi_{0.5}$ on \lib{}, \emph{``turn on the stove and put the moka pot on it''}: the VLA policy closes on the frying pan standing at the moka pot's usual location memorized in the original LIBERO training data; the evolved harness locates the named object, moka pot.}
\caption{\textbf{Evolved control routines fix concrete failure patterns.} Each row shows an episode under an earlier harness (starting with bare VLA policy) on the left, and under the revised \method{} harness on the right. Row (a) shows detecting and clearing arm jams, (b) synthesizing missing motion primitives, (c) rescaling contact forces, and (d) redirecting to target objects instead of memorized locations. \cref{fig:drivers1-more} shows more examples. See discussions on these revisions in \cref{sec:changes} and \cref{sec:appendix-qual}.}
 \label{fig:drivers1}
\end{figure}

\begin{figure}[t]
 \centering
 \includegraphics[width=\textwidth]{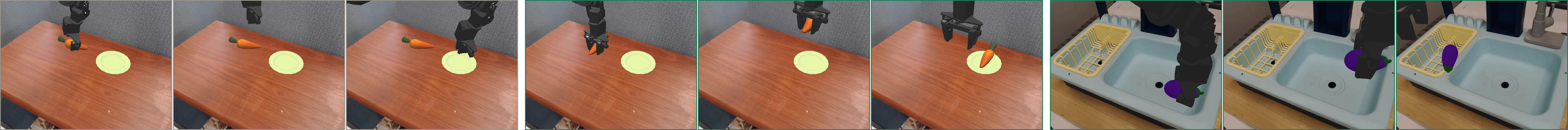}\\[1.5pt]
\panelcap{\textbf{(a)}~X-VLA on \widow{}, \emph{``put carrot on plate''} (left and middle) and \emph{``put eggplant into yellow basket''} (right): the memory records the fractions of object height that successfully lifted the object; closing at 18\% (raised to a 0.8\,cm minimum) on the carrot lifts nothing, while closing at 45\% of the object height (1.4\,cm on the carrot, 3.9\,cm on the eggplant) lifts both.}\\[5pt]
 \includegraphics[width=\textwidth]{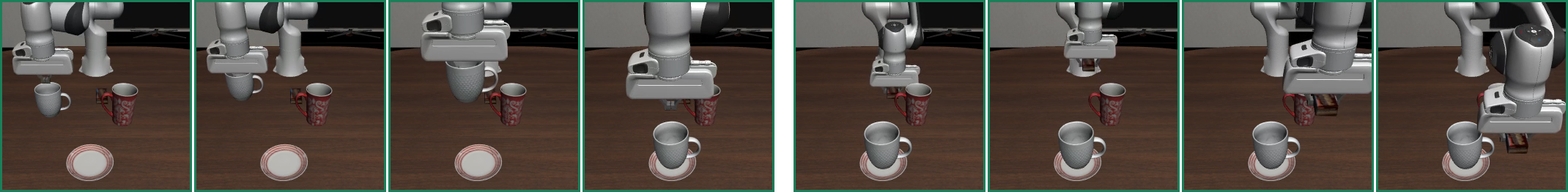}\\[1.5pt]
\panelcap{\textbf{(b)}~OpenVLA-7B on \lib{}, \emph{``put the white mug on the plate and put the chocolate pudding to the right of the plate''}: the memory records how often each grip style (pose) has worked for each object; the white mug is \textbf{spread from the inside} and stays upright, while the pudding box is \textbf{pinched}.}\\[5pt]
 \includegraphics[width=\textwidth]{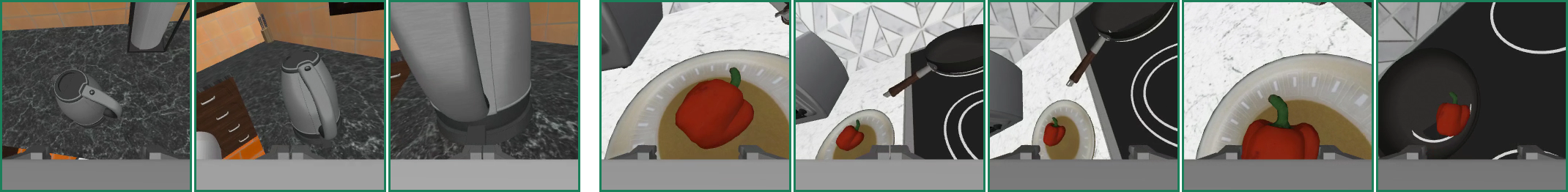}\\[1.5pt]
\panelcap{\textbf{(c)}~RLDX-1 on \casa{}, \emph{``\textbf{press} down the lever to turn on the electric kettle''} (left) and \emph{``\textbf{pick} the bell pepper from the plate and place it in the pan''} (right): the memory records that ``pressing'' tasks rarely \emph{hold} objects while ``picking'' tasks do; the harness keeps pressing on the lever with closed fingers, but backs off and retries when the fingers close on nothing over the pepper.}\\[5pt]
 \includegraphics[width=\textwidth]{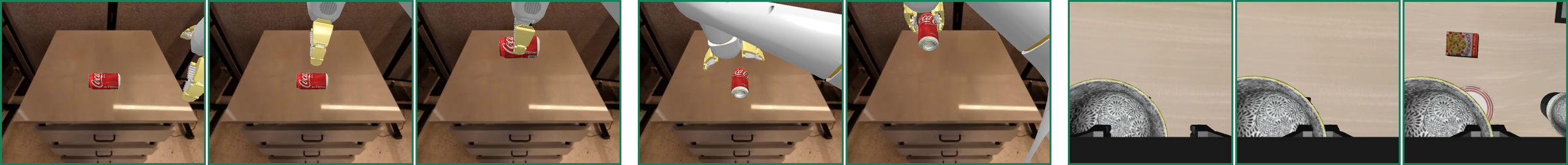}\\[1.5pt]
\panelcap{\textbf{(d)}~General X-VLA with the unified harness on \goog{}, \emph{``pick coke can''} (left and middle) and on LIBERO-Pro, \emph{``pick the akita black bowl on the cookie box and place it on the plate''} (right): the memory records the direction in which each robot's fingers close, and the harness turns the wrist accordingly; the Google robot closes its fingers across the can whether it lies sideways (left) or points toward the camera (middle), while the Franka arm grips the bowl's rim with one finger inside and one outside, following its correct wrist orientation.}
\caption{\textbf{Persistent memories emerge from harness evolution and inform action decisions.} Each row illustrates one form of memory mechanism; rows (a) and (b) show reusable execution experience, (c) task and environment context, and (d) embodiment-specific calibration. See discussions on these memories in \cref{sec:memory-results} and \cref{sec:appendix-qual}.}
 \label{fig:memory}
\end{figure}

\subsection{Harness Optimization Unlocks Substantial Robot Manipulation Headroom} \label{sec:headroom}

\para[para:specialist]{An evolved harness turns a general policy into a specialist.}
Without any weight updates, \method{} yields substantial gains across nearly all setups (\cref{tab:main}), and converts poorly performing cross-benchmark and out-of-domain policies (\cref{para:distance}) into functional specialists.
\cref{tab:main} compares harness design and finetuning directly: on \vlab{} under both protocols and on \widow{} under unseen scenes, \textbf{the evolved harness around the general policy outperforms the finetuned weights}, carrying the frozen policy past a full weight update; it falls short only on \widow{} under unseen tasks.
The gains \textbf{transfer strongly to unseen scenes and unseen tasks}: on \emph{unseen scenes}, they generalize across novel scene arrangements, with the only minor exception of RLDX-1 on \casa{}, which improved only on its optimization episodes (\cref{fig:traj-scenes}); every setup also transfers positively to \emph{unseen tasks} (\cref{tab:main}).
This generalization reflects the structure of the revisions themselves: control routines work from \textbf{measured geometry and the robot's own sensing}, not from static object names or positions written into the code (\cref{sec:changes}), and memories are often indexed by \textbf{measured physical dimensions} (\cref{sec:memory-results}).
These results show that the domain gap traditionally closed through demonstration collection~\citep{kim2024openvla,intelligence2025pi_} can also be closed through automated harness design.

\para[para:balance]{The harness dynamically balances policy and programmatic control.}
In several setups, the evolved harness bypasses the frozen policy entirely on benchmarks a general policy was never trained for.
For example, the general X-VLA on \widow{} commands a fully open gripper at every step, so the harness learns to \textbf{disregard the policy entirely} and evolves into a vision-based code-as-policy agent (\cref{fig:drivers2}i; details in \cref{sec:appendix-qual}).
On its optimization episodes, this harness raised success from 0\% to 100\% (\cref{fig:traj-tasks,fig:traj-scenes}), and achieved 87.5\% on unseen scenes and 66.7\% on unseen tasks.
In contrast, the unified harness evolved on finetuned policies (\cref{para:unified}) \textbf{keeps the VLA specialists in charge and intervenes only when they fail}.
As shown in \cref{fig:unified-drivers}a, if late in an episode the gripper under the VLA policy's control has never approached the instructed object, the harness takes over via the pointing tool, and directly commands the arm to reach and grasp it. 
The harness also learns to hand control back to the VLA policy if it is already closing in on the object (\cref{fig:unified-drivers}b).
Most evolved harnesses settle between these two ends: \textbf{the policy retains the motion it executes well, while the code takes over the decisions the policy gets wrong}, such as lifting a jammed arm clear before re-querying the policy (\cref{fig:drivers1}a), or carrying the gripper to the named object before the policy grasps it (\cref{fig:drivers1}d).
Crucially, we never instructed the coding model how to divide control; each harness arrived at its structure from rollout evidence alone.

\para[para:unified]{A unified harness works across robots.}
We next test whether a single harness can generalize across embodiments (\cref{para:protocols}; results in \cref{tab:unified}, full evolution trajectories in \cref{fig:traj-unified}).
The general X-VLA model scores near zero across all benchmarks, as it is entirely out of distribution (\cref{tab:unified}).
The harness then evolves into a pure code-as-policy agent, replacing the general VLA entirely (as in \cref{para:balance}). 
It raises every benchmark from (near) zero, reaching 41.6\% on the training pool and 28.8\% on held-out evaluation.
This unified harness reads everything that differs between robots at runtime from the action specification and proprioception, never hardcoding embodiment-specific constants (\cref{sec:changes,sec:memory-results}). It thus \textbf{operates across embodiments without domain-specific finetuning}.
Using finetuned specialists as base policies, the unified harness yields smaller gains. These policies already perform well (58.0\% training, 45.9\% held-out), and so our auto-evolved harness \textbf{correctly preserves their native competence}.
As such, the unified harness strikes the same balance as individual setups (\cref{para:balance}): it evolves into a pure code-as-policy agent when wrapping the general VLA, but correctly preserves native competence when wrapping finetuned specialists.

\para[para:iterative]{Iterative refinement further improves the automatically designed harness.}
A coding model can already design an agent from scratch in one attempt.
But the agent itself generates new evidence through its own execution, and the same optimization process of \method{} can use that evidence to refine it further.
The harness after the first iteration (\emph{One-Shot} in \cref{tab:main,tab:unified}) represents a \textbf{one-shot agent design}: the coding model sees the bare VLA policy's rollout failures (iteration 0) and produces an agent in a single attempt (iteration 1). 
Subsequent iterations observe the behavior of this agent and refine its control logic based on newly exposed failure modes (\cref{para:optimizationloop}), up to iteration 5 (\emph{Iterative} in \cref{tab:main,tab:unified}).
\cref{fig:traj-scenes,fig:traj-tasks,fig:traj-unified} illustrate these optimization trajectories, showing \textbf{steady, mostly monotonic progression} over iterations.
As shown in \cref{tab:main}, the iteratively evolved harness outperforms the one-shot design on the training set in every setup (\cref{fig:traj-scenes,fig:traj-tasks}) and on held-out sets, except RLDX-1 on \casa{} under unseen scenes.
For example, the code-as-policy agent on \vlab{} is built in its first attempt, and only its own rollouts reveal subtler failures, which later revisions repair (\cref{fig:drivers2}g,h; details in \cref{sec:appendix-qual}).
\textbf{Iterative refinement also improves generalization}: while some one-shot designs fall below the bare VLA policy on unseen scenes or tasks, the iteratively evolved harness surpasses it on nearly every held-out set (\cref{tab:main}).
The improvement is not limited to adding new tools: later revisions also refine the division of control between the policy and programmatic logic, as discussed in \cref{para:balance}.
These results show that the harness is not merely automatically designed, but can \textbf{iteratively self-improve} by inspecting the execution traces of its own previous designs.

\subsection{Evolved Control Routines Driving the Performance Gains} \label{sec:changes}

The evolved harnesses address concrete failure modes with targeted control routines.
\cref{fig:drivers1,fig:drivers2} show representative revisions across benchmarks.
The harnesses learn to \textbf{detect and clear arm jams} that the policy cannot recover from (\cref{fig:drivers1}a), and to \textbf{synthesize missing motion primitives}, such as a side approach under a shelf board or a handle pull on a cabinet (\cref{fig:drivers1}b, \cref{fig:drivers1-more}e).
On \casa{}, they \textbf{rescale contact forces} when the policy pushes a door too gently (\cref{fig:drivers1}c), and \textbf{tether the mobile base} to active workspaces when it drifts away (\cref{fig:drivers1-more}f).
Other routines make perception more reliable: the harness redirects the arm to the target object when a policy overfit to its training scenes reaches for a memorized location (\cref{fig:drivers1}d), and \textbf{adapts to perception limitations} by working down ranked candidates and checking for firm grasps from depth (\cref{fig:drivers2}g,h).
The unified harness also \textbf{adapts to each robot's setup}, such as its speed: it ends each move when the arm arrives, not on a fixed time budget (\cref{fig:unified-drivers}c), which contributed to raising the training-pool success rate on the slower Google robot from 11.0\% to 30.2\%.
\textbf{\cref{sec:appendix-qual} details each routine.}

\subsection{Persistent Memory Emerges from Harness Evolution} \label{sec:memory-results}

Across nearly all setups, \method{}'s evolved harnesses develop persistent memory structures that carry experience across episodes.
\cref{tab:memory} lists examples of memory entries and how often each held-out episode reads the memory.
Rather than following any predefined memory schema, \method{} discovers diverse forms of memories that capture \textbf{reusable execution experience, task and environment context, and embodiment-specific calibration}.

\begin{wraptable}{R}{0.58\textwidth}
 \centering
 \caption{\textbf{Auto-designed persistent memories are frequently accessed and yield highly useful information to guide later actions.} \emph{Reads per Episode}: average frequency of memory reads in held-out episodes. \emph{Actionable Reads}: the share of these reads that find stored info.}
 \label{tab:memory}
 \scriptsize
 \setlength{\tabcolsep}{2pt}
 \renewcommand{\arraystretch}{1.1}
 \def\rowline{\noalign{{\color{black!15}\hrule height 0.3pt}}}
 \begin{tabular}{@{}>{\raggedright}p{0.2625\linewidth}>{\raggedright\arraybackslash}p{0.4317\linewidth}rr@{}}
 \toprule
Setting & Memory Entries & \shortstack{Reads per\\Episode} & \shortstack{Actionable\\Reads} \\
 \midrule
X-VLA on\newline VLABench & finger depths that have been lifting objects & 3.4 & 98\% \\
 \rowline
X-VLA on\newline SimplerEnv WidowX & height at which to close the fingers, per measured object height & 3.5 & 96\% \\
 \rowline
MolmoAct-2 on\newline LIBERO-Pro-10 & average depth of successful grasps below an object's top & 9.9 & 99\% \\
 \rowline
$\pi_{0.5}$ on\newline RoboCerebra & scene-reset period of this simulation & 13.0 & 95\% \\
 \rowline
RLDX-1 on\newline RoboCasa-365 & task instructions that need a grip & 11.0 & 92\% \\
 \bottomrule
 \end{tabular}
\end{wraptable}
Memories \textbf{learn from the harness's own grasps}: one records the \emph{grasp heights} that lifted objects, relative to each object's measured height, and consults them on new objects (\cref{fig:memory}a); another remembers the \emph{grip style} (\emph{pinch} or \emph{spread}) that held each object (\cref{fig:memory}b).
Indexed by measured physical properties or object identity, this experience carries over to novel objects and new tasks.
Memories also track \textbf{implicit task and environment clues}, such as which instructions need \emph{holding} objects (\cref{fig:memory}c) and the 150-step scene-reset period of \cere{}.
The unified harness \textbf{calibrates itself to each robot}, recording which way the robot's fingers close to orient later grasps (\cref{fig:memory}d).
\textbf{\cref{sec:appendix-qual} details each type of memory}.

Importantly, these memories are \textbf{data structures within the designed robot agent}, not the coding model's own context: they accumulate information during the agent's rollouts and persist across episodes at execution time.
As shown in \cref{tab:memory}, the harnesses \textbf{actively consult their memory} several times per episode on average, and \textbf{over 90\% of these reads are actionable}, that is, they successfully find a stored entry, even on held-out episodes unseen during harness optimization.
Because every evaluation starts with an empty memory to be accumulated, these entries' designs effectively carry over to held-out episodes and remain highly useful.

\section{Conclusion} \label{sec:conclusion}

We introduced \method{} to automate the design and optimization of agent harnesses around frozen robot policies. 
Across diverse benchmarks and frozen policies, the evolved harnesses substantially improve performance on held-out scenes and tasks.
Even when the policy is untuned for the target environment, \method{} can adapt it to a specialist in out-of-domain settings, and can even outperform its finetuned version.
Crucially, the harness continues to self-improve beyond its initial design, using the behavior of previous versions as new evidence for subsequent refinements.
Emergent from harness optimization, we observe intuitive control routines that resolve specific failure modes, and persistent memories that carry experience across episodes to inform later decisions.
\method{} can also evolve a unified harness that works across multiple embodiments, adapting itself to each robot.
Together, these results suggest that substantial headroom in robot manipulation resides in the agent harness, and that this headroom can be unlocked by automatically evolving the harness for novel setups.

\section{Limitations} \label{sec:limitations}

\para{Scaling cap of iterations.}
As shown in \cref{fig:traj-scenes,fig:traj-tasks,fig:traj-unified}, harness performance on the training episodes improves consistently across iterations.
However, this work uses a fixed maximum of 5 optimization iterations (\cref{para:config}).
More iterations may yield further gains, but the point at which returns saturate remains to be explored.
Nevertheless, even with a moderate budget of only 5 iterations (\cref{para:cost}), \method{} achieves substantial gains that transfer to held-out settings.

\para{More complex manipulation.}
A natural next step is extending \method{} to tasks with richer contact dynamics, such as dexterous in-hand manipulation or tool use, where the space of failure modes is larger than in pick-and-place.
The range of control routines and memory structures that automated evolution can discover in these settings is an interesting open question.

\para{Sim-to-real transfer.}
All evaluations run in simulation. 
Because our harness reads only the standard input space (\emph{e.g.,} camera images, depth and the robot's own readings), its interface should carry over to a real robot.
Testing our simulation-evolved harnesses on real robots and investigating the number of further improvement iterations on a real deployment are future work.

\bibliography{main}
\bibliographystyle{main}

\appendix
\renewcommand{\topfraction}{0.95}
\renewcommand{\bottomfraction}{0.8}
\renewcommand{\textfraction}{0.05}
\renewcommand{\floatpagefraction}{0.8}
\setcounter{topnumber}{3}
\setcounter{bottomnumber}{2}
\setcounter{totalnumber}{4}
\setlength{\textfloatsep}{12pt plus 2pt minus 2pt}
\setlength{\floatsep}{8pt plus 2pt minus 2pt}
\setlength{\intextsep}{8pt plus 2pt minus 2pt}

\section{Evolved Control Routines and Persistent Memories in Detail} \label{sec:appendix-qual}

This appendix details \method{}'s auto-evolved control routines of \cref{sec:changes} and the persistent memories of \cref{sec:memory-results}; \cref{fig:drivers1-more,fig:drivers2,fig:unified-drivers} show more of the revisions.

\subsection{Control Routines}

\para{Detecting and clearing arm jams.}
When the arm is jammed against an object, the bare policy may continue issuing similar commands because its visual observations remain nearly unchanged during a jam, so it tips objects over and cannot recover.
The evolved harness learns to monitor discrepancies between commanded and actual gripper movements; when disagreement persists, it lifts the arm clear before re-querying the policy (\cref{fig:drivers1}a).

\begin{figure}[t]
 \centering
 \includegraphics[width=\textwidth]{figs/head_optimization.pdf}\\[2pt]
 \includegraphics[width=\textwidth]{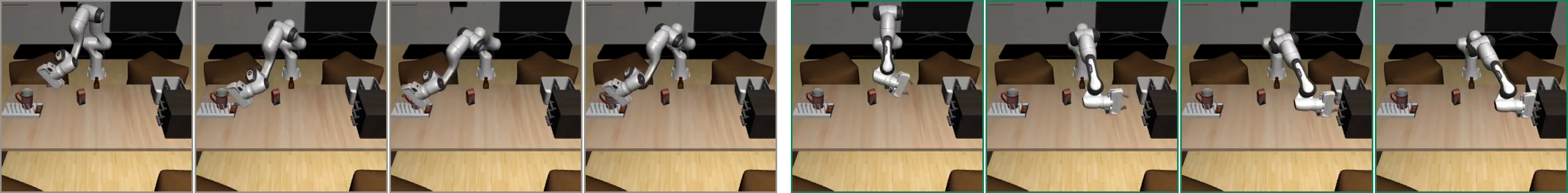}\\[1.5pt]
\panelcap{\textbf{(e)}~SmolVLA on \cere{}, \emph{``Open the wooden cabinet top region''}: the VLA policy fails entirely, missing required motions; the evolved harness correctly locates the cabinet, closes on the handle and pulls the drawer out.}\\[5pt]
 \includegraphics[width=\textwidth]{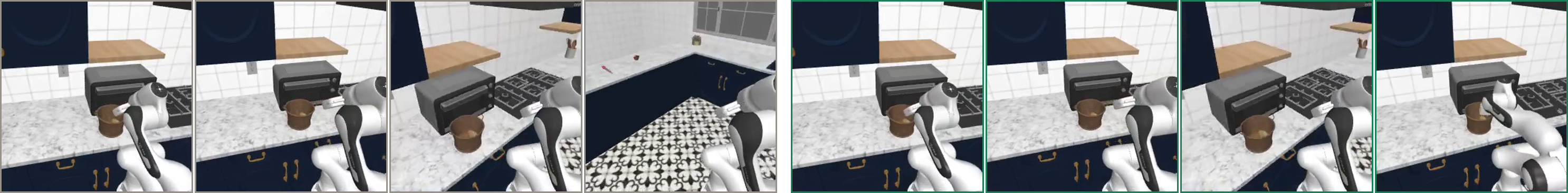}\\[1.5pt]
\panelcap{\textbf{(f)}~RLDX-1 on \casa{}, \emph{``Open the cabinet, pick up the bread from the cabinet and place it in the basket. Then move the basket to the dining counter.''}: the VLA policy drifts the base away and never returns to the counter; the evolved harness detects the drift and retraces back to the counter.}
\caption{\textbf{More evolved control routines.} Rows read as in \cref{fig:drivers1}. Row (e) shows synthesizing missing motion primitives, and (f) tethering mobile bases to active workspaces.}
 \label{fig:drivers1-more}
\end{figure}

\para{Synthesizing missing motion primitives.}
Some tasks require motion primitives absent from the base policy or the earlier harness, such as approaching an object from the side rather than from above.
For retrieving a book from the shelf on \vlab{}, the earlier harness defaults to top-down motion and collides with the overhead board. The evolved harness detects overhead obstructions via depth maps and computes horizontal approach vectors to reach beneath them (\cref{fig:drivers1}b).
The same pattern appears elsewhere: on \cere{}, the harness adds a wrist-rotation primitive for pouring with $\pi_{0.5}$, while for cabinet opening with SmolVLA, it locates the cabinet with the pointing model, finds its handles in the depth image, then aligns and pulls (\cref{fig:drivers1-more}e).

\para{Rescaling contact forces.}
On \casa{}, some failures arise from \textbf{insufficient contact force} rather than trajectory misalignment.
For example, the toaster-oven door requires closure within 0.5\% of its opening range, but the policy RLDX-1 pushed too gently. The evolved harness rescales action magnitudes while preserving direction, providing sufficient force to latch the door (\cref{fig:drivers1}c).

\para{Tethering mobile bases to active workspaces.}
On \casa{}, the harness also learns to constrain \textbf{mobile-base drift} around active workspaces.
It defines workspace anchors as locations where the last gripper activations occurred, gives the base a driving budget, and retraces the base to the latest anchor once the budget is spent (\cref{fig:drivers1-more}f).

\para{Redirecting to target objects instead of memorized locations.}
For spatially shifted scenes such as LIBERO-Pro, policies overfit on original LIBERO consistently reach toward wrong items at memorized locations~\citep{zhou2025libero}. \method{} independently diagnosed this memorization, and evolved the harness to query the pointing model for the target item. With RLinf-$\pi_{0.5}$, it learns to locate the instructed moka pot rather than reaching to its memorized position (\cref{fig:drivers1}d).

\begin{figure}[t]
 \centering
 \includegraphics[width=\textwidth]{figs/head_optimization.pdf}\\[2pt]
 \includegraphics[width=\textwidth]{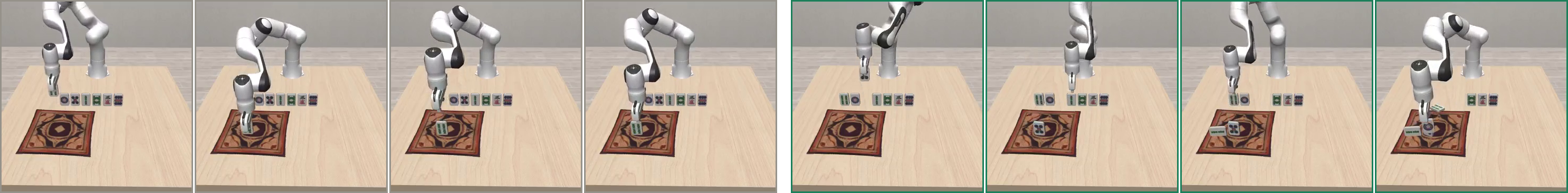}\\[1.5pt]
\panelcap{\textbf{(g)}~X-VLA on \vlab{}, \emph{``Pick up the mahjong of 1 pin''}: the earlier harness commits to just a single tile; the evolved harness works down its ranked candidate list of multiple tiles.}\\[5pt]
 \includegraphics[width=\textwidth]{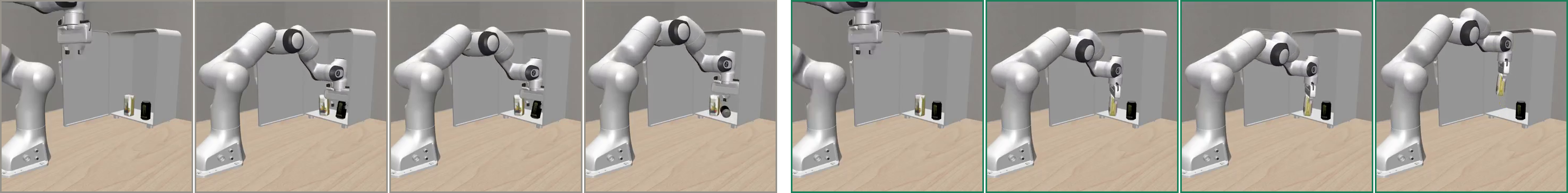}\\[1.5pt]
\panelcap{\textbf{(h)}~X-VLA on \vlab{}, \emph{``It's a hot day and I could use some refreshment''}: the earlier harness judges its grip empty after lifting the drink and puts it down; the evolved harness verifies grasps from the depth image under the fingertips and keeps hold of the drink.}\\[5pt]
 \includegraphics[width=\textwidth]{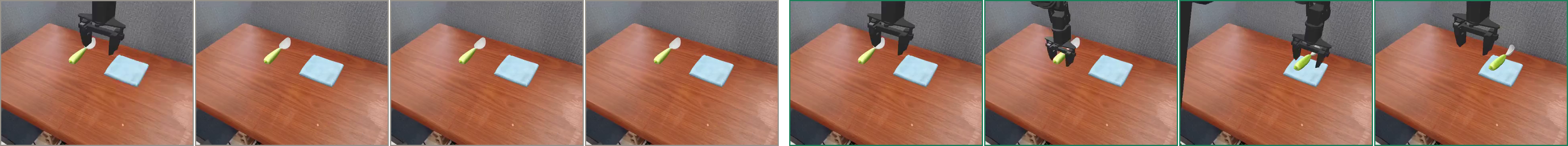}\\[1.5pt]
\panelcap{\textbf{(i)}~X-VLA on \widow{}, \emph{``put the spoon on the towel''}: the VLA policy (un-tuned and out of distribution) fails entirely; \method{} replaces it with a code-as-policy agent that picks and places the spoon correctly.}
\caption{\textbf{Evolved harnesses also enhance the robotic agent's perception precision and reliability.} Panels read as in \cref{fig:drivers1}, discussions in \cref{sec:changes}. Rows (g) and (h) show adapting to perception limitations, and (i) replacing an unusable policy with a code-as-policy agent (\cref{para:balance}).}
 \label{fig:drivers2}
\end{figure}

\para{Replacing an unusable VLA policy.}
The base (un-finetuned) X-VLA on \widow{} commanded a fully open gripper at every step, producing no useful control signal.
\method{} learned to \textbf{disregard the policy entirely}, as its commands barely change with the robot's observations.
Instead, the harness evolved into a complete vision-based picking agent, akin to code-as-policy~\citep{liang2023code,tsui2026free,elmaaroufi2026rho}: parsing instructions, detecting target objects via open-vocabulary detectors, estimating 3D poses from depth maps, computing grasp vectors, and executing trajectories directly (\cref{fig:drivers2}i).

\para{Adapting to perception limitations.}
When the VLA policy is unusable, the harness evolves into a pure code-as-policy agent (\cref{para:balance}). Crucially, such agents continue to \textbf{refine their own perception and control mechanisms} across iterations (\cref{para:iterative}).
On \vlab{}, the vision tools could not reliably distinguish mahjong tile faces. Rather than relying on a single vision prediction, the harness maintains a ranked candidate list, sequentially interacting with candidate tiles until the task completes (\cref{fig:drivers2}g).
Vision-based grasp verification could also be unreliable, causing the agent to prematurely put down objects it was holding. The harness thus replaces \textbf{visual grasp checks} with \textbf{a depth check under the fingertips}, verifying grasps by the presence of a surface just below the fingers (\cref{fig:drivers2}h).

\begin{figure}[t]
 \centering
 \includegraphics[width=\textwidth]{figs/head_optimization.pdf}\\[2pt]
 \includegraphics[width=\textwidth]{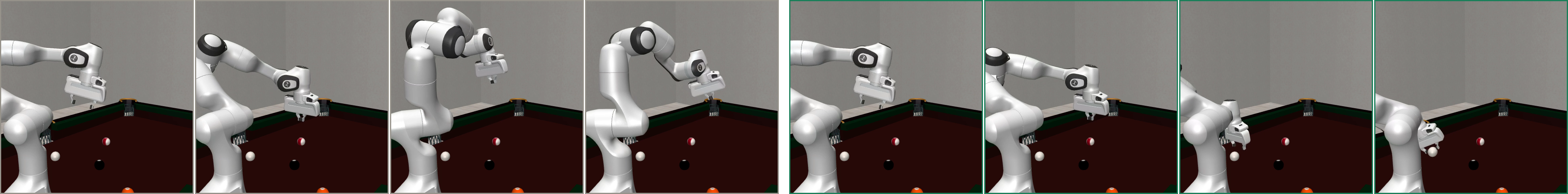}\\[1.5pt]
\panelcap{\textbf{(a)}~Finetuned X-VLA on \vlab{}, \emph{``Please put the white in any hole.''}: the VLA policy sweeps above the table and never descends; the revised harness borrows the arm late in the episode and brings the gripper down to the ball.}\\[5pt]
 \includegraphics[width=\textwidth]{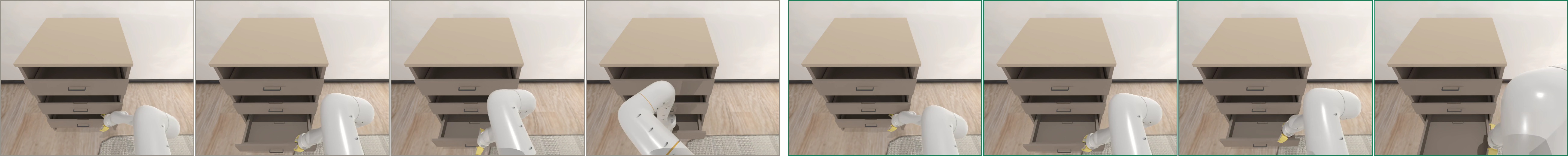}\\[1.5pt]
\panelcap{\textbf{(b)}~Finetuned X-VLA on \goog{}, \emph{``open bottom drawer''}: the earlier harness takes the arm over after the drawer is already open and pushes it shut again; the revised harness no longer takes the arm from a policy that has already reached its target, and the drawer stays open.}\\[5pt]
 \includegraphics[width=\textwidth]{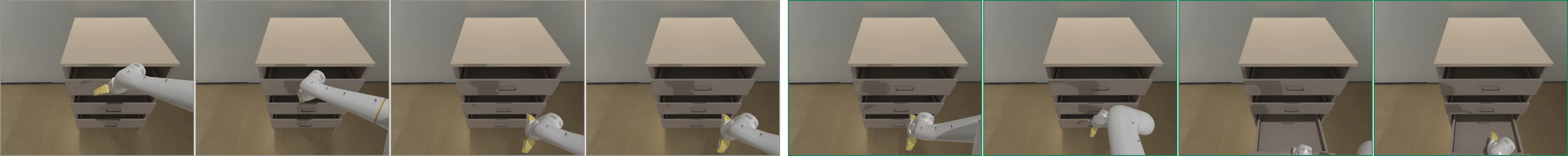}\\[1.5pt]
\panelcap{\textbf{(c)}~General X-VLA on \goog{}, \emph{``open bottom drawer''}: the earlier harness presses the hand onto the drawer front and times out on a clock set for a faster robot; the revised harness ends moves on arrival and reaches the handle from the front along the surface normal.}
\caption{\textbf{The unified harness learns to take over the arm command from a failing policy and to leave a progressing one alone, and calibrates itself to each robot's configurations.} Each row shows an episode under the harness before (left) and after (right) an iteration's revision.}
 \label{fig:unified-drivers}
\end{figure}

\para{Per-robot control-rate independence.}
The unified harness (\cref{para:unified}) evolves mechanisms that adapt control to each robot across platforms (see \cref{fig:unified-drivers}).
Each motion phase used to end on a fixed time budget, which on a slower platform (\emph{e.g.,} the 3\,Hz Google robot, \cref{fig:unified-drivers}c) expires while the arm is still in transit.
The evolved harness ends a move only when the arm arrives or stops making progress, measured by the distance between the arm's position and its target rather than by elapsed time, and reaches handles on vertical surfaces from the front along the measured surface normal. The revision that introduced these changes raised the agent's performance on the Google robot from 11.0\% to 30.2\% on the training pool.

\subsection{Persistent Memories}

\para{Learning execution parameters from experience.}
The harnesses store empirical execution statistics that inform later control decisions.
On \widow{}, the harness records the grasp heights that have successfully lifted an object, as a function of measured object height, so that effective grasp parameters transfer to novel objects (\cref{fig:memory}a).
On \lib{}, the MolmoAct-2 harness instead averages the depth of successful grasps below the object top, as a share of the object's height, and uses this statistic to parameterize fallback grasps.
The OpenVLA-7B harness similarly records the grip style, \emph{pinch} or \emph{spread}, that succeeds for each object, so future grasps select the proven pose when the object's shape allows both (\cref{fig:memory}b).
Crucially, across these designs, memory is indexed by \textbf{actual, measured physical properties or object identity} rather than specific task names, so experience generalizes to novel objects across different tasks.

\para{Tracking implicit task and environment clues.}
Persistent memory also captures recurring clues in the task instruction and environment.
On \casa{}, the harness associates instruction verbs with the need to \emph{hold (``pick up'')} an object: once sufficient evidence accumulates for a \emph{``picking''} task, it enables an \emph{empty-grasp recovery routine} that retries when fingers close on nothing, while leaving the routine disabled for \emph{``pressing''} tasks where a closed empty gripper is intentional (\cref{fig:memory}c).
On \cere{}, the harness discovers and stores the environment's 150-step scene-reset period, allowing later episodes to anticipate resets and take over a placement early enough to finish it before the reset.
Memories are also used to cache inference outputs, such as VLM/LLM parses of multi-step instructions, so that repeated executions avoid redundant computation.

\para{Persistent self-calibration across embodiments.}
In the unified harness (\cref{para:unified}), persistent memory additionally stores embodiment-specific calibration.
Because finger motion differs across robots, the harness opens and closes the empty gripper, observes the resulting motion in the depth image, and \textbf{stores the kinematic mapping in memory}.
This calibration guides grasp orientation across different embodiments (\emph{e.g.,} on the Franka arm and on the Google robot) (\cref{fig:memory}d).
The same routine records the resting finger configuration, providing a reference for detecting empty grasps on robots reporting finger positions.

\section{Implementation Details} \label{sec:appendix-impl}

\para[para:appendix-bench]{Benchmarks.}
\lib{} evaluates long-horizon task sequences from LIBERO~\citep{liu2023libero} with initial object locations swapped while preserving original natural language instructions. These spatial swaps follow the LIBERO-Pro benchmark protocol~\citep{zhou2025libero} designed to distinguish genuine manipulation skill from spatial memorization via overfitting on LIBERO.
\cere{}~\citep{han2025robocerebra} tests on a sequence of short manipulation tasks with scene resets every 150 control steps, challenging the ability to handle sudden environment changes.
\vlab{}~\citep{zhang2025vlabench} requires identifying specific target items among visually similar distractors with a focus on visual grounding.
\casa{}~\citep{nasiriany2024robocasa,nasiriany2026robocasa365} places a mobile manipulator in varied kitchen environments, requiring coordinated mobile navigation and arm control.
\widow{}~\citep{li2024evaluating} deploys a WidowX manipulator within photorealistic scenes, testing cross-embodiment transfer under visual domain shift.
\goog{}~\citep{li2024evaluating} deploys the Google mobile manipulator under the same visual domain shift as \widow{}, and is included in the test bed of the unified harness (\cref{para:protocols}).

\begin{figure}[t]
 \centering
 \includegraphics[width=\textwidth]{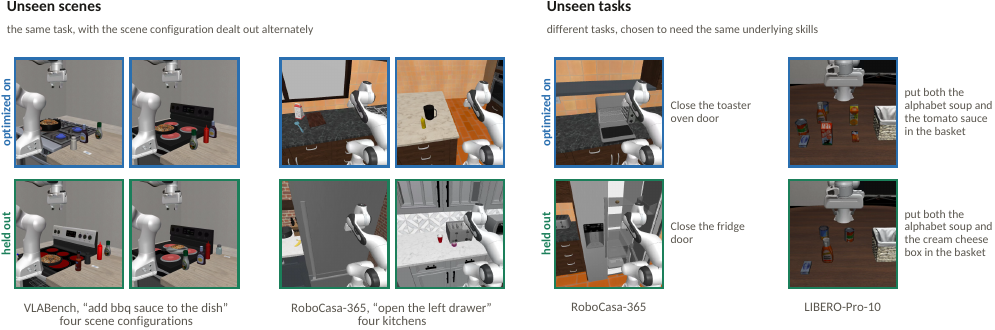}
\caption{\textbf{Held-out episodes differ from the optimization episodes either in scene arrangement (\emph{unseen scenes}) or in the task itself (\emph{unseen tasks})} (\cref{para:protocols}). \emph{Left, unseen scenes:} the same tasks with scene configurations dealt out alternately. \emph{Right, unseen tasks:} different tasks that require similar underlying skills.}
 \label{fig:splits}

 \vspace{14pt}

 \includegraphics[width=\textwidth]{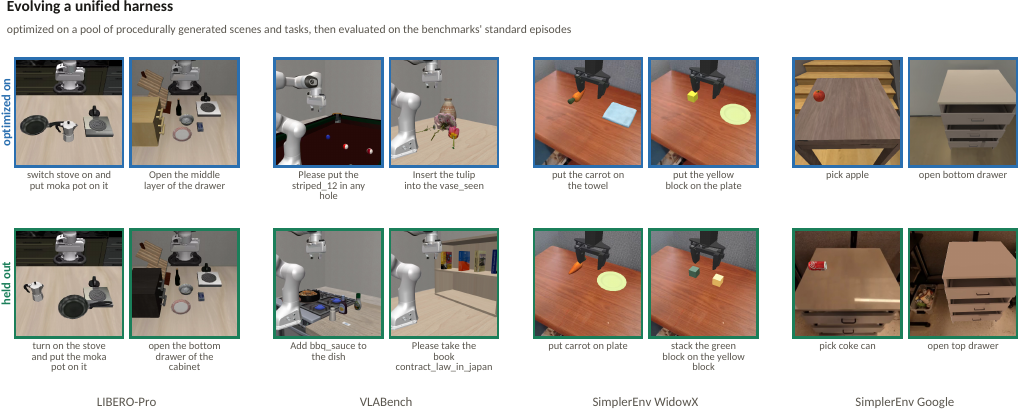}
\caption{\textbf{The unified harness is evolved only on a training pool of procedurally generated episodes and evaluated on the benchmarks' own standard episodes} (\cref{para:protocols,para:unified}). \emph{Top, optimized on:} procedurally generated scenes, objects, instructions \& tasks form the training pool. \emph{Bottom, held out:} the benchmarks' own standard episodes, unseen during harness evolution.}
 \label{fig:unified-pool}
\end{figure}

\para[para:appendix-episodes]{Episode splits.}
\cref{fig:splits} shows the split of a benchmark's episodes under the \emph{unseen scenes} and \emph{unseen tasks} protocols, and \cref{fig:unified-pool} shows the training pool and the held-out episodes of the unified harness (\cref{para:protocols}).
\cref{tab:tasksplit} lists, for each benchmark, the tasks used for the harness evolution and the held-out tasks under the \emph{unseen tasks} protocol. The held-out tasks share underlying manipulation skills with the harness optimization tasks. The task split of a benchmark is shared by all the policies evaluated on it (\cref{tab:main}). Every task is run for the same number of trials (\emph{i.e.,} ``episodes''), as listed next to each benchmark.
The unified harness (\cref{para:unified}) is evolved on a training pool of 2{,}201 procedurally generated episodes, and evaluated on the 2{,}160 standard episodes.

\begin{table}[t]
 \centering
\caption{\textbf{Optimization and held-out tasks under the \emph{unseen tasks} protocol.} On \cere{} a task is a chain of sub-task instructions stated here in brief. On \vlab{} and \casa{} the instruction of a task changes with the sampled objects. \emph{Trials} is the number of trials (\emph{i.e.,} episodes) per task.}
 \label{tab:tasksplit}
 \scriptsize
 \setlength{\tabcolsep}{3pt}
 \renewcommand{\arraystretch}{1.15}
 \begin{tabular}{@{}p{0.11\linewidth}rp{0.40\linewidth}p{0.40\linewidth}@{}}
 \toprule
Benchmark & Trials & Optimization Tasks & Held-Out Tasks \\
 \midrule
\textcolor{bcliberopro10swap}{\textbf{LIBERO-Pro-10}} & 10 & \emph{turn on the stove and put the moka pot on it}; \newline \emph{put the black bowl in the bottom drawer of the cabinet and close it}; \newline \emph{put both the alphabet soup and the tomato sauce in the basket}; \newline \emph{put both the cream cheese box and the butter in the basket}; \newline \emph{put the white mug on the left plate and put the yellow and white mug on the right plate} & \emph{put the yellow and white mug in the microwave and close it}; \newline \emph{put both moka pots on the stove}; \newline \emph{put both the alphabet soup and the cream cheese box in the basket}; \newline \emph{put the white mug on the plate and put the chocolate pudding to the right of the plate} \\
 \midrule
\textcolor{bcrobocerebrapublic60}{\textbf{RoboCerebra}} & 10 & cream cheese, popcorn and butter, each into a white storage box; \newline cream cheese into a white storage box, then the frypan onto the rack; \newline plate, popcorn and cookies rearranged on the dining set, then wine poured onto the plate; \newline chocolate pudding from the cabinet top onto the plate, then wine poured onto the plate; \newline chocolate pudding and butter into a black bowl, the stove turned on, then the frypan onto the stove & cookies into a basket, tomato sauce and bbq sauce poured onto the frypan, wine poured into a mug, then another mug into a wooden tray; \newline cream cheese and popcorn into a white bowl, then the frypan onto the cabinet top; \newline the short cabinet opened, butter into the cabinet, then tomato sauce poured onto the plate; \newline the short cabinet opened, cream cheese into the cabinet, chocolate pudding into a black bowl, then wine poured into a mug; \newline white bowl from the cabinet top, cookies and chocolate pudding into the bowl, the microwave opened, then the bowl into the microwave \\
 \midrule
\textcolor{bcvlabenchxvlatracks14}{\textbf{VLABench}} & 10 & select the named fruit; \newline select the named mahjong tile; \newline select the named poker card & select the named drink out of a fridge; \newline select the $n$-th largest poker card; \newline select the mahjong tile of the unique type \\
 \midrule
\textcolor{bcsimplerwidowxvm}{\textbf{SimplerEnv WidowX}} & 24 & \emph{put the spoon on the towel}; \newline \emph{stack the green block on the yellow block} & \emph{put carrot on plate}; \newline \emph{put eggplant into yellow basket} \\
 \midrule
\textcolor{bcrobocasa365target}{\textbf{RoboCasa-365}} & 50 & close the toaster oven door; \newline open the cabinet door; \newline open the drawer; \newline pick an object from the counter and place it in the cabinet; \newline press the start button on the microwave & close the fridge door; \newline pick an object from the plate and place it in the pan on the stove; \newline fully slide the top dishwasher rack out; \newline press down the lever to turn on the electric kettle \\
 \bottomrule
 \end{tabular}
\end{table}

\para[para:cost]{Cost of automated harness optimization.}
A complete evolution of 5 iterations (\cref{para:config}) takes under 6 hours of coding-model time and stays under \$150 on average, with no demonstrations collected and no model weights updated. Through \method{}, the engineering loop that is slow by hand (\cref{sec:intro}) becomes an inexpensive routine easily repeatable for every new robot setup.

For more details, refer to our codebase at \url{https://github.com/roboarchitect-ai/RoboArchitect}.

\section{Harness Evolution Trajectories} \label{sec:appendix-traj}

\begin{figure}[!h]
 \centering
 \includegraphics[width=0.95\textwidth]{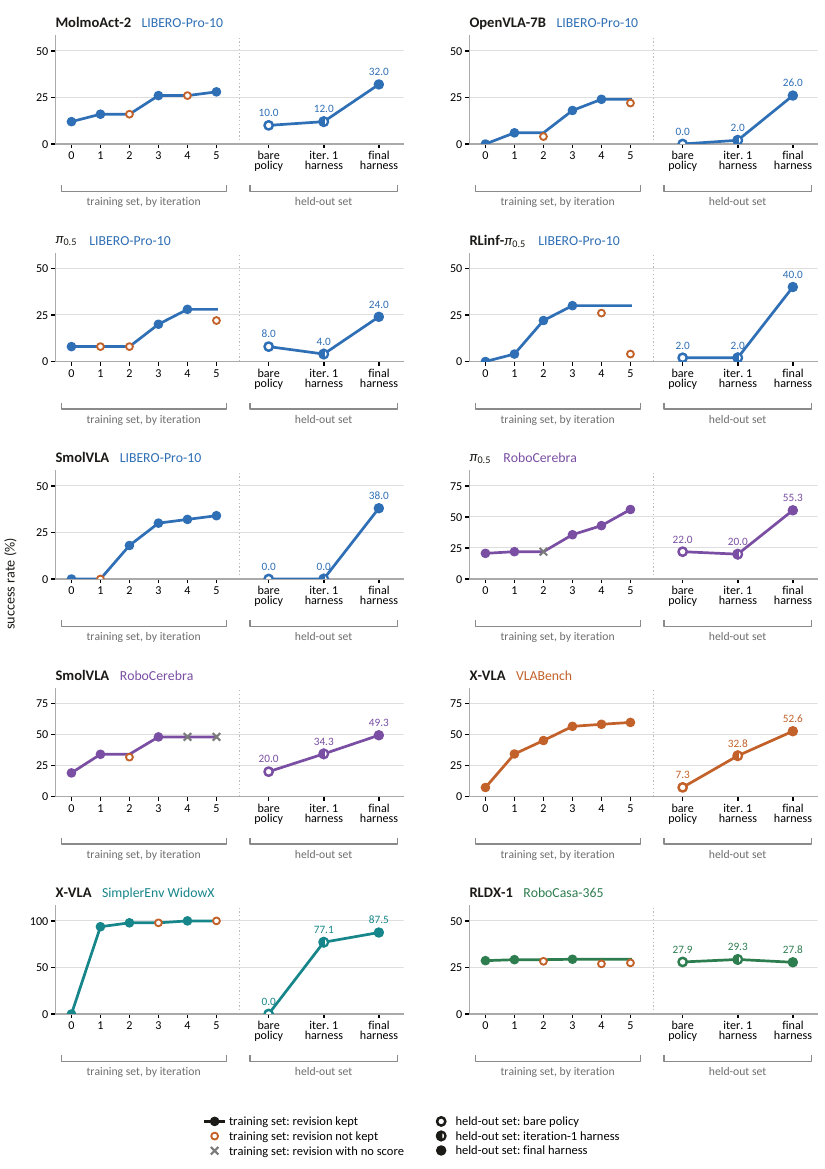}
\caption{\textbf{Under \emph{unseen scenes}, the harness improves over iterations on the training set, and the final harness outperforms the one-shot harness on nearly every held-out set.} Left of the dotted line is the training episode set, iteration by iteration; iteration 0 is the bare VLA policy, filled markers denote accepted harness revisions that improved task success over prior iterations, and open markers denote revisions that were not accepted. Right of the dotted line are the scores on the held-out set: the bare policy, the harness after iteration 1 (a single-attempt agent design) and the final iteratively refined harness. The held-out set evaluates identical tasks under different scene arrangements (\cref{para:protocols}).}
 \label{fig:traj-scenes}
\end{figure}

\begin{figure}[htbp]
 \centering
 \includegraphics[width=0.95\textwidth]{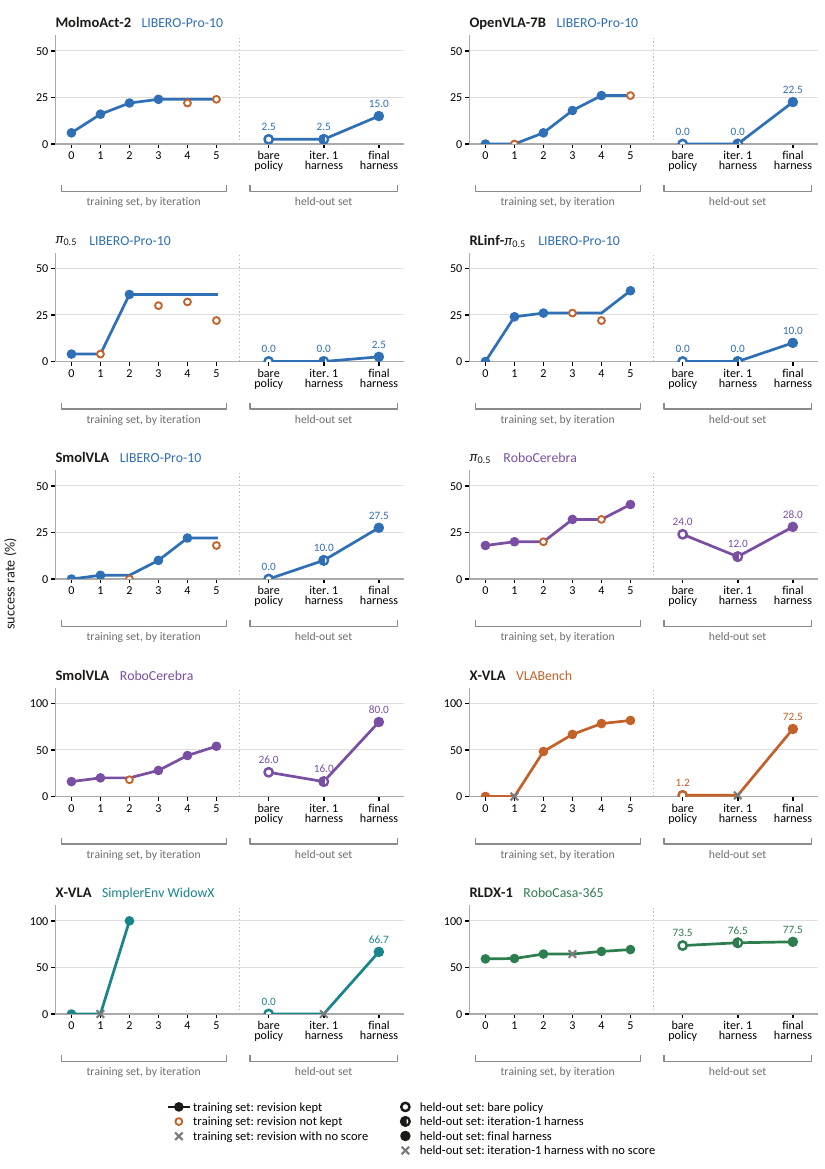}
\caption{\textbf{Under \emph{unseen tasks}, the harness improves over iterations on the training set, and the final harness outperforms the bare VLA policy and the one-shot harness on every held-out set.} Labels and panels read as in \cref{fig:traj-scenes}. A cross marks a revision that could not be evaluated; when this is the first revision, no single-attempt harness is available for held-out evaluation. The held-out set evaluates unseen tasks requiring shared underlying skills (\cref{para:protocols}).}
 \label{fig:traj-tasks}
\end{figure}

\begin{figure}[htbp]
 \centering
 \includegraphics[width=0.95\textwidth]{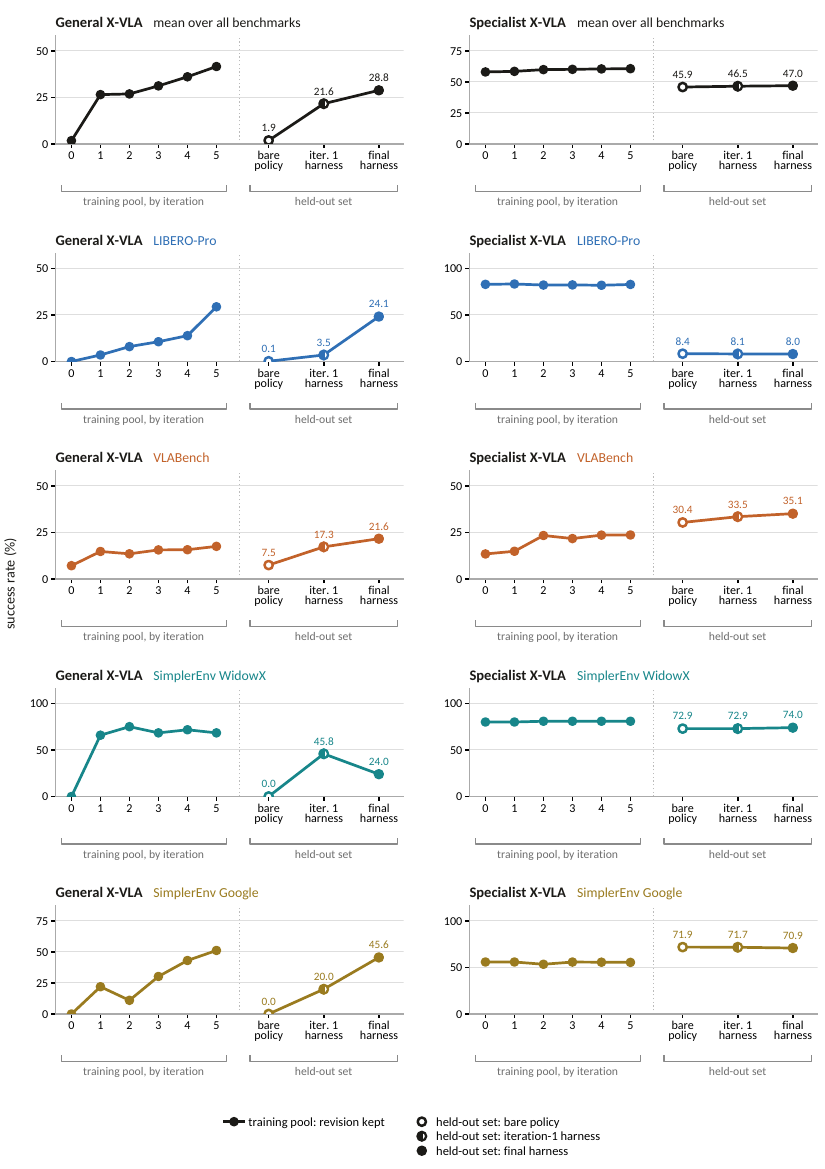}
\caption{\textbf{Around the general X-VLA, the unified harness improves on every benchmark of the training pool over iterations; around the finetuned specialists, it leaves their already competent scores intact.} Left column, using the general X-VLA checkpoint as policy; right column, using benchmark-finetuned X-VLA specialists as policy. The first row is the mean over the benchmarks and each later row is one benchmark. Left of the dotted line shows the performance on the procedurally generated training pool, iteration by iteration (iteration 0 is the bare policy; filled markers denote accepted harness revisions). Right of it are the scores on the benchmark's standard episodes (unseen in the training pool): the bare policy, the harness after iteration 1 (a single-attempt agent design) and the final iteratively refined harness.}
 \label{fig:traj-unified}
\end{figure}

\cref{fig:traj-scenes,fig:traj-tasks,fig:traj-unified} show the complete optimization update chain for every experimental setting under both held-out evaluation protocols (\cref{para:protocols}) and the unified agent design (\cref{para:unified}).

\end{document}